\documentclass[10pt,leqno]{amsart}
\usepackage{graphicx}
\usepackage{indentfirst,csquotes}

\usepackage{amssymb,amsthm,amsmath}
\usepackage{xcolor,hyperref}

\hypersetup{colorlinks=true, linkcolor=black, filecolor=black, urlcolor=black }

\begin{document}

\title{Extending Tabular Denoising Diffusion Probabilistic Models for Time-Series Data Generation}

\author{Umang Dobhal}
\author{Christina Garcia}
\author{Sozo Inoue}

\address{Kyushu Institute of Technology, Fukuoka, Japan}

\email{dobhal-umang638@mail.kyutech.jp}
\email{alvarez7.christina@gmail.com}
\email{sozo@brain.kyutech.ac.jp}

\date{\today}
\maketitle

\begin{abstract}
Diffusion models are increasingly being utilised to create synthetic tabular and time series data for privacy-preserving augmentation. Tabular Denoising Diffusion Probabilistic Models (TabDDPM) generate high-quality synthetic data from heterogeneous tabular datasets but assume independence between samples, limiting their applicability to time-series domains where temporal dependencies are critical. To address this, we propose a temporal extension of TabDDPM, introducing sequence awareness through the use of lightweight temporal adapters and context-aware embedding modules. By reformulating sensor data into windowed sequences and explicitly modeling temporal context via timestep embeddings, conditional activity labels, and observed/missing masks, our approach enables the generation of temporally coherent synthetic sequences. Compared to baseline and interpolation techniques, validation using bigram transition matrices and autocorrelation analysis shows enhanced temporal realism, diversity, and coherence. On the WISDM accelerometer dataset, the suggested system produces synthetic time-series that closely resemble real world sensor patterns and achieves comparable classification performance (macro F1-score 0.64, accuracy 0.71). This is especially advantageous for minority class representation and preserving statistical alignment with real distributions. These developments demonstrate that diffusion based models provide effective and adaptable solutions for sequential data synthesis when they are equipped for temporal reasoning. Future work will explore scaling to longer sequences and integrating stronger temporal architectures.
\end{abstract}

\bigskip

\section{Introduction}
\label{section:Introduction}

Due to applications in activity analytics, context-aware services, and health monitoring, Human Activity Recognition (HAR) employing wearable and smartphone sensors has been thoroughly researched. In particular, triaxial accelerometers offer an inexpensive and convenient way to record human motion, allowing for large scale behavioral modeling and assistive technologies \cite{wisdm2010}. But gathering sizable, varied, and accurately labeled time-series datasets for HAR is still difficult. The procedure is costly, time consuming, and frequently limited by participant recruiting restrictions and privacy laws, which limit the diversity and accessibility of training data for downstream models \cite{wisdm2010}.

Recent developments in deep generative modeling have created new avenues for resolving privacy concerns and data shortages. In particular, diffusion based methods have proven to be remarkably effective in producing diverse and realistic synthetic samples in a variety of domains, including time-series and tabular data \cite{tabddpm2023} \cite{tabsyn2024} \cite{forestdiffusion2024}. Through denoising procedures, these models gradually convert random noise into structured data, attaining high fidelity and controllability. For mixed type datasets, extensions like latent diffusion and hybrid VAE diffusion frameworks have further enhanced sample efficiency and scalability \cite{tabsyn2024} \cite{timeldm2024} \cite{timeautodiff2024}. Additionally, to better retain temporal dynamics in generated sequences, domain specific modifications have started incorporating interpretable decompositions (such as trend and seasonality) and temporal structure \cite{diffusiontts2024}.

Applying diffusion based generative models to multimodal sensor data like accelerometer signals for HAR presents a number of outstanding issues despite these encouraging advancements. Specifically, models need to: 
\begin{itemize}
    \item manage heterogeneous data features and class imbalance 
    \item maintain computational efficiency while producing high fidelity synthetic signals and
    \item preserve fine-grained temporal relationships across numerous sensor channels. 
\end{itemize}
To guarantee the usefulness of synthetic HAR data for model training and assessment, several factors must be taken into consideration.

In this study, we modify the TabDDPM framework \cite{tabddpm2023}, which was initially created for tabular data, to facilitate the efficient and realistic creation of multivariate time-series signals. The objective is to generate temporally coherent synthetic human activity data that may be applied to downstream recognition tasks. In order to accomplish this, we add temporal embeddings, conditional context, and missing value awareness to TabDDPM, which keeps its computational simplicity and flexibility while successfully capturing sequential dependencies in smartphone accelerometer data.

The main contributions of this study are as follows:
\begin{enumerate}
    \item We point out the main shortcomings of current generating methods for HAR data and emphasize the necessity of creating synthetic data that is computationally efficient, interpretable, and temporally consistent.
    \item To enable robust synthesis for multivariate accelerometer time-series, we suggest a diffusion based generative framework improved with diffusion timestep embeddings, side conditional embeddings, and an observed value mask.
    \item We show that the suggested method preserves signal diversity and dynamics across activity classes, providing a workable remedy for data imbalance and scarcity in HAR research.
\end{enumerate}

In extensive experiments on the WISDM dataset, the proposed method generates synthetic time-series data with a macro F1-score of 0.64 and accuracy of 0.71, matching baseline and SMOTE methods while showing improved temporal coherence versus original TabDDPM, as confirmed by bigram and autocorrelation metrics. Bigram transition matrices show that proposed TabDDPM maintains sharp and realistic step-by-step transitions that closely match real sensor signals, while autocorrelation analysis demonstrates that it faithfully preserves both short range stochastic behaviour and mild periodic dependencies found in real sequences. These enhancements produce synthetic data with improved temporal structure and variability.

The remainder of this paper is organized as follows. Section~\ref{section: Related Literatures} reviews existing literature on generative models for tabular and time-series data. Section~\ref{section: Methodology} presents the proposed framework in detail. Section~\ref{section: Results} discusses experimental results and evaluation metrics, followed by discussion in Section~\ref{section: Discussion} and concluded with the conclusion and future direction in Section~\ref{section: Conclusion and Future Directions}.

\section{Related Literature}
\label{section: Related Literatures}

\subsection{Challenges in Data Collection}
The effectiveness of both discriminative and generative models is hampered by the inherent difficulty of gathering data for human activity recognition (HAR) utilizing wearable or smartphone sensors. The majority of publicly accessible datasets, including WISDM \cite{wisdm2010}, are collected in controlled laboratory settings where participants carry out predetermined tasks in brief, isolated sessions. Although these settings guarantee accurate measurements, they do not capture natural variability in movement patterns, transitions, and environmental contexts and lack the ecological validity of real world behavior \cite{chan2024capture}.

Annotation complexity presents a second significant obstacle. It is necessary to accurately segment and label continuous accelerometer data, frequently using heuristic based segmentation or manual supervision, which results in labeling noise and discrepancies. The scalability of HAR datasets is greatly limited by this laborious and error prone technique \cite{nguyen2024sok}.

Lastly, distributional changes between data sources are brought about by sensor and device heterogeneity. Cross-user or cross-device learning is complicated by domain disparities caused by differences in device orientation, sample rate, and location (e.g., phone in pocket vs. hand). The creation of deep generative models, such as diffusion based architectures, which need huge, diverse, and temporally consistent samples to accurately describe human motion, is hampered by the combination of these problems, which restrict the volume, balance, and realism of accessible data.

Recent research has also demonstrated how data imbalance and scarcity directly impair activity detection ability in real-world gesture-based and healthcare contexts. To address the severe class imbalance in nurse activity recognition, for instance, synthetic skeleton data generation using large language models has been investigated. This is because short-duration and fine-grained sub-activities produce disproportionately few samples and noisy pose extractions \cite{dobhal2024synthetic}. Similarly, to compensate for limited and uneven temporal samples, diffusion-based synthetic generation with structured post-processing has been studied for gesture phase recognition \cite{dobhal2025sample}. These studies show that real-world sensing environments frequently have fragmented, noisy, and class-skewed data distributions beyond laboratory HAR benchmarks. This highlights the need for principled synthetic data generation strategies to improve diversity, balance, and statistical fidelity while maintaining activity semantics.

\subsection{Limitations of Generative Approaches}
Despite a great progress in generative modeling for structured data, existing frameworks face critical challenges when applied to time-series sensor data such as those used in human activity recognition (HAR).

While early generative models, such as GANs and VAEs, were effective in simulating low dimensional data distributions, they had trouble with the temporal dependencies included in multivariate sensor signals. Although a supervised embedding network was used to add temporal links in TimeGAN \cite{timegan2019}, training instability and mode collapse remained major problems, especially for long or irregular sequences. While VAE-based techniques like VAEM \cite{vaem2020} enhanced feature variety for heterogeneous data, they were not designed to maintain the inter-timestep correlations necessary for motion continuity.

Approaches like TabDDPM \cite{tabddpm2023} and TabSyn \cite{tabsyn2024} demonstrate strong performance for tabular data generation with the advent of diffusion probabilistic models. Nevertheless, these models lost the dynamic aspects of human motion because they were made for static features rather than sequential inputs. Similar to this, TimeAutoDiff \cite{timeautodiff2024} and TimeLDM \cite{timeldm2024} extended diffusion to temporal data, but they are still computationally demanding and require numerous reverse diffusion stages for high-fidelity reconstruction, which makes them unsuitable for real time applications.

Heterogeneous feature representation, which is essential for HAR data including both continuous (accelerometer readings) and categorical (activity labels, user IDs) variables, is not sufficiently addressed by the majority of current frameworks. Models frequently lack interpretability when it comes to how synthetic sequences maintain activity semantics or feature correlations, even with latent diffusion techniques \cite{timeldm2024}.

In summary, previous generative methods face three key obstacles:
\begin{itemize}
    \item \textbf{Temporal discontinuity} involving weak modeling of sequential dependencies across time.
    \item \textbf{High computational cost} due to excessive diffusion steps for realistic synthesis.
    \item \textbf{Low interpretability} with limited understanding of activity specific latent structures.
\end{itemize}

These shortcomings collectively highlight the need for a lightweight, temporally consistent, and interpretable diffusion framework motivating the proposed modifications described in this study.

\subsection{Limitation of TabDDPM}
\begin{figure}[h]
    \centering
    \includegraphics[width=1\linewidth]{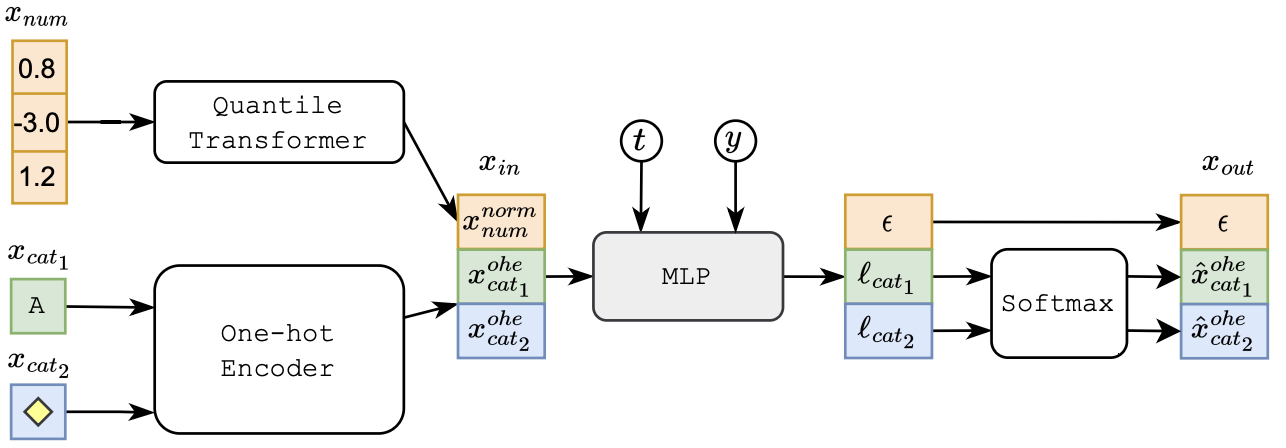}
    \caption{TabDDPM architecture for tabular data with quantile-transformed numerical and one-hot encoded categorical features.}
    \label{fig: architecture}
\end{figure}

Although TabDDPM \cite{tabddpm2023} brought diffusion probabilistic modeling to the tabular realm with remarkable outcomes, its design is still essentially limited for time-series applications. The model ignores temporal dependencies that are crucial in sequential data, like accelerometer signals, and instead assumes independent and identically distributed characteristics. Its architecture as shown in Figure \ref{fig: architecture} uses one-hot encoding for categorical attributes and quantile transformation for numerical data, followed by an MLP-based noise prediction network. This feature-wise approach works well for static data, but it ignores cross timestep relationships, which are crucial for simulating dynamics between successive sensor readings. As a result, produced samples could match general feature distributions but not the sequential flow seen in actual time-series.

In previous work, we investigated these limitations by adapting TabDDPM for time-series gesture data through feature-level temporal encoding and a post-processing selection strategy \cite{dobhal2025sample}. Because TabDDPM does not inherently guarantee that generated samples are temporally or semantically consistent, a filtering mechanism based on Mahalanobis distance and classifier confidence was introduced to remove out-of-distribution and low-fidelity synthetic samples. While this improved statistical alignment and class-wise performance, the reliance on post-generation filtering highlighted that temporal coherence was not natively modeled within the diffusion architecture itself.

Additionally, TabDDPM lacks contextual conditioning methods, which makes it unable to include outside data like activity type or user-specific context. Its robustness in practical sensor contexts, where signal loss or irregular sampling is widespread, is limited by the lack of missing data handling. Furthermore, even while the model's computational architecture is effective for tabular data, it does not scale well for lengthy temporal sequences that call for frame-wise noise estimation.

Overall, even though TabDDPM provides a robust and computationally efficient basis for diffusion-based data synthesis on tabular data, it is still unsuitable for multi-sensor, temporally structured datasets. In order to enable coherent, condition-aware time-series creation, our architecture builds on this robustness by incorporating temporal adapters and contextual embeddings.

\subsection{Overview of Proposed Method}
While TabDDPM \cite{tabddpm2023} adequately represents static tabular data, it lacks tools for handling temporal dependencies, conditional control, and missing values prevalent in sensor based HAR datasets. To fill these limitations, we augment the diffusion framework with temporal and context-aware components that increase sequential coherence, class-conditioned synthesis, and robustness to incomplete data. This adaptation is intended to allow diffusion-based models to better capture temporal structure in accelerometer time-series, serving as a design framework that addresses key limitations of tabular diffusion approaches when applied to time-dependent signal modeling.
\section{Methodology}
\label{section: Methodology}

\subsection{Proposed Architecture}
The suggested framework expands upon TabDDPM \cite{tabddpm2023}, modifying it to produce multivariate sensor time-series data. Although TabDDPM works well for static tabular data, it lacks the explicit temporal modeling and conditioning techniques necessary for accelerometer sequences including human action. Our architecture provides enhanced sequential coherence, conditional controllability, and robustness to missing values by adding three context-aware embedding blocks and temporal adapters while keeping the fundamental preprocessing and diffusion framework of TabDDPM.

\vspace{-0.5em}
\subsubsection{Architecture Overview}
\begin{figure}
    \centering
    \includegraphics[width=1\linewidth]{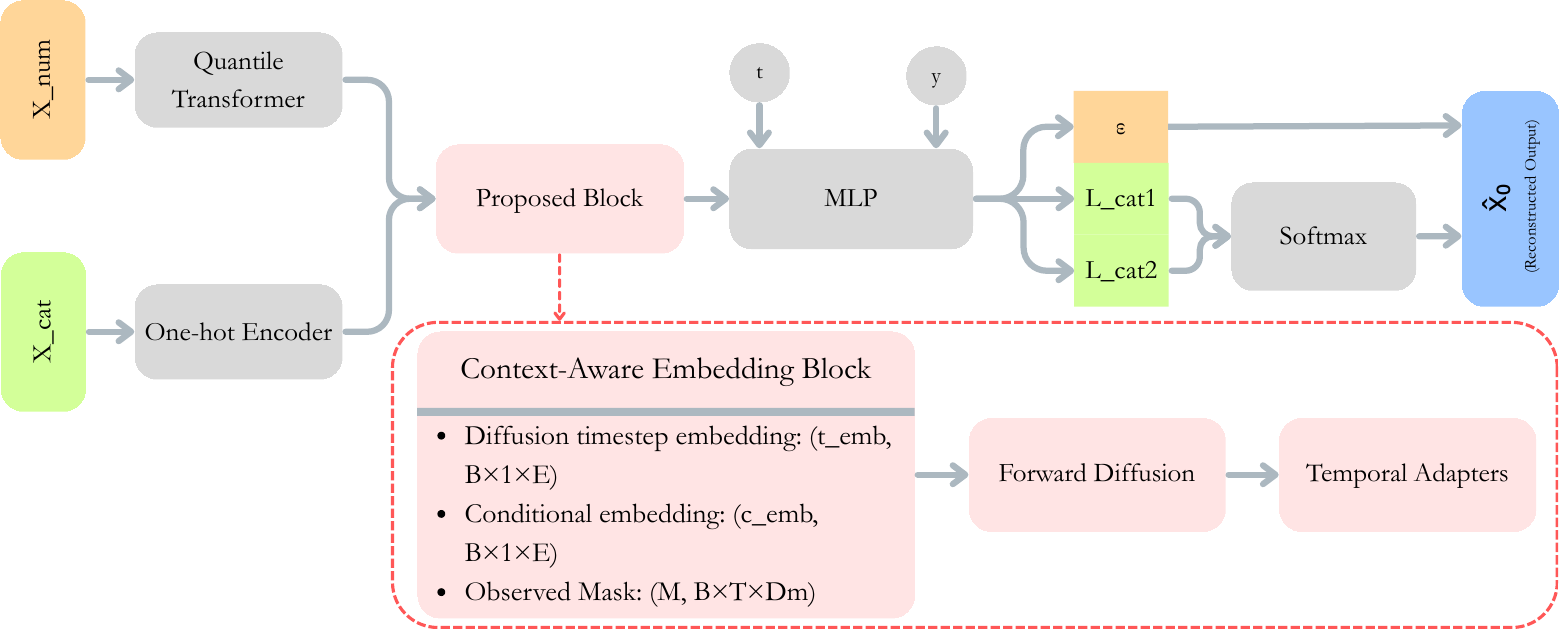}
    \caption{Proposed Temporal TabDDPM architecture with context-aware embeddings and temporal adapters.}
    \label{fig: proposed_architecture}
\end{figure}

Figure \ref{fig: proposed_architecture} shows the overall pipeline: input sensor features are first normalised via a quantile transformer for numerical channels and one-hot encoded for categorical attributes, replicating the TabDDPM preprocessing. These per feature encodings are concatenated into $x_0 \in \mathbb{R}^{B \times T \times D}$. Before entering the forward diffusion block, three additional modules – Timestep Embedding, Conditional Embedding, and Observed/Missing Mask – are merged with $x_0$ to form an augmented latent. The forward diffusion then progresses as $q(x_t | x_{t-1})$, followed by Temporal Adapters and an MLP noise predictor, culminating in reconstruction of the generated sequence.

\vspace{-0.5em}
\subsubsection{Context-Aware Embedding Block}
\label{Section: Context Aware}
\begin{enumerate}
    \item Timestep Embedding (t\_emb): To encode temporal context into the denoising network, each diffusion step \textit{t} is embedded into a learnable vector. This makes it possible for the model to differentiate between early and late denoising phases, which improves its ability to replicate the changing dynamics of human motion.
    \item Conditional Embedding (c\_emb): Side information, like activity labels or user identities, is encoded by an auxiliary embedding layer. This allows for direct control over the kind of activity being synthesized by supporting class-conditioned sequence generation.
    \item Observed/Missing Mask (M): The time-series signal's observed and missing elements are indicated by a binary mask. This makes the model resilient to missing sensor data and uneven sampling, which are prevalent in real world HAR datasets.
\end{enumerate}

These three embeddings are concatenated with $x_0$ to yield a unified latent representation that feeds into the diffusion process, thereby enhancing TabDDPM with temporal, conditional, and missing data awareness.

\vspace{-0.5em}
\subsubsection{Temporal Adapters}
\label{section:temporal_adapters}

After the forward diffusion block, the latent representation passes through \textit{Temporal Adapters}, lightweight modules inserted inline with the denoising network. These adapters capture short range temporal dependencies without requiring full transformer or recurrent architectures. Following prior work on parameter efficient adapters~\cite{houlsby2019adapter, pfeiffer2021adapterfusion, ruoss2023diffusionadapters}, the modules introduce only a small number of trainable parameters while substantially improving temporal coherence.

The adapter adopts a compact Conv1D based structure designed to model local temporal patterns efficiently. Given an input sequence \[x \in \mathbb{R}^{B \times T \times D},\] the adapter performs the following computations.

\textbf{Temporal Position Encoding.}
    Standard sinusoidal positional embeddings~\cite{ho2020denoising} are added to encode timestep information:
    \begin{equation}
        x' = x + \mathrm{PE}(t)
    \end{equation}
    where $\mathrm{PE}(\cdot)$ denotes sinusoidal positional encodings.

\textbf{Temporal Convolution.}
    Two lightweight 1D convolutions operate along the temporal dimension:
    \begin{equation}
        h_t = \mathrm{Conv1D}_{k=3}\!\left(x'_t\right), \qquad
        h   = \mathrm{Dropout}\!\left(\mathrm{ReLU}(h_t)\right),
        \label{eq:adapter_conv}
    \end{equation}
    providing local receptive fields and efficient parameter sharing across timesteps.

\textbf{Feature Projection.}
    The temporal features are projected back to the original dimensionality:
    \begin{equation}
        \mathrm{Adapter}(x) = Wh + b,
        \label{eq:adapter_linear}
    \end{equation}
    where $W \in \mathbb{R}^{D \times D}$ is a learnable linear projection.

The resulting adapter output is passed to the TabDDPM MLP denoiser. This compact design increases temporal expressiveness at an $O(T)$ computational cost significantly lower than self-attention mechanisms with $O(T^2)$ complexity while preserving the simplicity of the original TabDDPM architecture.

\subsubsection{Integration with the Diffusion Process}
The forward diffusion framework from TabDDPM \cite{tabddpm2023} is extended with conditional and temporal context through Temporal Adapters and conditional embeddings, enabling more effective sequence recovery in the reverse process. This design elevates the denoising MLP, supporting high fidelity sensor time-series generation with robust handling of missing values and realistic temporal patterns.

Key improvements over the baseline:
\begin{itemize}
    \item Temporal coherence via temporal adapters and timestep embeddings.
    \item Class conditional synthesis of activity specific sequences.
    \item Robustness to missing or uneven sampling using an observation mask.
\end{itemize}

These enhancements enable more realistic and controllable human activity sequence generation for data augmentation and HAR training.

\subsection{Dataset}
\label{subsection: dataset}

This study uses the WISDM (Wireless Sensor Data Mining) dataset~\cite{wisdm2010}, a widely used benchmark for smartphone based Human Activity Recognition. The dataset contains triaxial accelerometer readings sampled at 20\,Hz while participants perform six activities: walking, jogging, sitting, standing, upstairs, and downstairs. Each record consists of $(x, y, z)$ acceleration values, a timestamp, and a user identifier.

To reduce user specific variability and ensure computational feasibility, data from the first five participants (out of twenty) were selected. The attributes included in the dataset are summarized in Table~\ref{tab:attributes}.

\begin{table}[h]
\centering
\caption{Dataset Attributes Description}
\label{tab:attributes}
\begin{tabular}{|l|l|l|}
\hline
\textbf{Attribute} & \textbf{Description} & \textbf{Type} \\
\hline
user          & User identifier               & Integer \\
activity      & Activity label (6 classes)     & Categorical \\
x, y, z       & Accelerometer readings (m/s²)  & Float \\
timestamp\_ms & Timestamp in milliseconds      & Integer \\
\hline
\end{tabular}
\end{table}

\begin{figure}[h]
    \centering
    \includegraphics[width=0.75\linewidth]{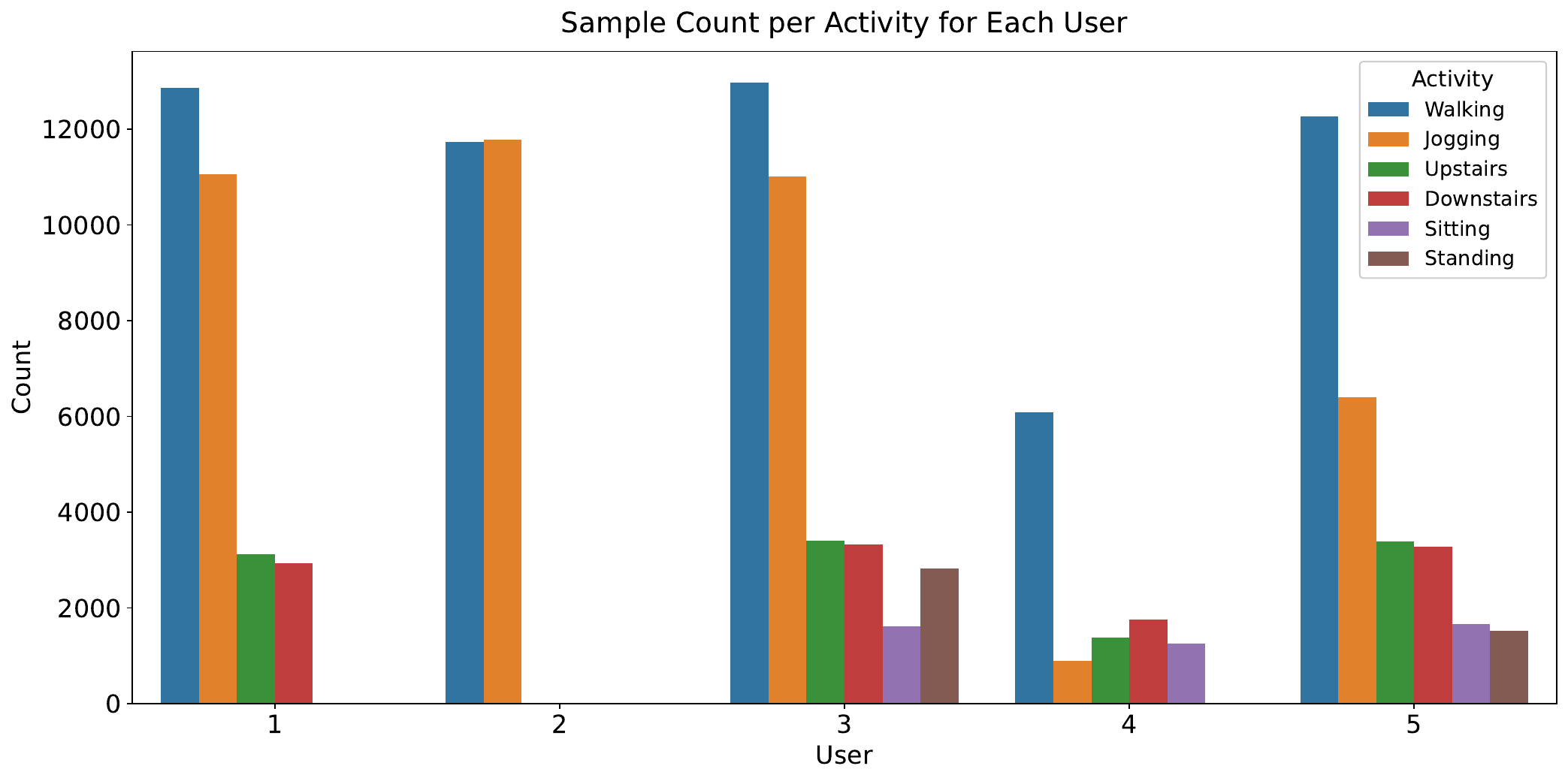}
    \caption{Activity distribution across the five selected users.}
    \label{fig: activity distribution}
\end{figure}

Figure~\ref{fig: activity distribution} shows the activity distribution across the selected users and highlights the natural imbalance between locomotion based and sedentary activities.

An 80/20 stratified split was applied to create training and test sets, ensuring proportional representation of all activity classes. The held out test set was used exclusively for evaluating both the generative model and downstream classification performance.

\subsection{Data Preprocessing}
\label{section: preprocessing}

Raw accelerometer data were segmented into overlapping windows of 100 samples ($T=100$), each labeled with the dominant activity. Window overlap was set to 50 samples to increase sample count and preserve temporal continuity~\cite{xaviar2023robust}. Only activity labels were provided to the generative framework.

Numeric features ($x, y, z$) were normalized using a Quantile Transformer fitted on the training set and reused during sampling to ensure consistent scaling. Categorical attributes were integer encoded, and user identity was excluded from the generative process. Missing values were handled using a binary mask $M \in \{0,1\}^{B \times T \times D}$, distinguishing observed from imputed entries and improving robustness to data gaps~\cite{xaviar2023robust}. Data splits followed the 80/20 train-test proportion described in Section~\ref{subsection: dataset}, with the training set further split 80/20 for model validation. The final model inputs were tensors of shape $(N, 100, 3)$ paired with their corresponding activity label and mask, forming the triplet $(x_0, y, M)$ used throughout all diffusion steps.

\vspace{-0.5em}
\subsection{Model Training}
\label{section: model training}

The proposed TabDDPM was trained to learn the conditional distribution of multivariate sensor sequences through a denoising diffusion process~\cite{ho2020denoising, nichol2021improved}. The model predicts the Gaussian noise added at each diffusion step.

\vspace{-0.8em}

\subsubsection{Training Configuration}
The model was implemented in PyTorch and trained on the preprocessed WISDM data using a batch size of 64, learning rate of $1\times10^{-3}$, weight decay of $1\times10^{-5}$, and the AdamW optimizer. A cosine learning rate schedule and gradient clipping (threshold $1.0$) were applied. The diffusion process used 1000 timesteps with a cosine noise schedule $\{\beta_t\}_{t=1}^{1000}$.

At each step, noise $\epsilon \sim \mathcal{N}(0, I)$ was added to $x_0$ to obtain $x_t$, and the denoising network $\epsilon_\theta(x_t, t, y, M)$ was trained using the standard DDPM objective:

\begin{equation}
\mathcal{L}_{simple} = \mathbb{E}_{x_0, \epsilon, t} \left[ \| \epsilon - \epsilon_\theta(x_t, t, y, M) \|_2^2 \right],
\end{equation}

where $y$ is the activity label and $M \in \{0,1\}^{B \times T \times D}$ is the observed value mask concatenated with $x_t$ before embedding.

\vspace{-1em}

\subsubsection{Optimization and Stability}
\begin{figure}[h]
    \centering
    \includegraphics[width=0.8\linewidth]{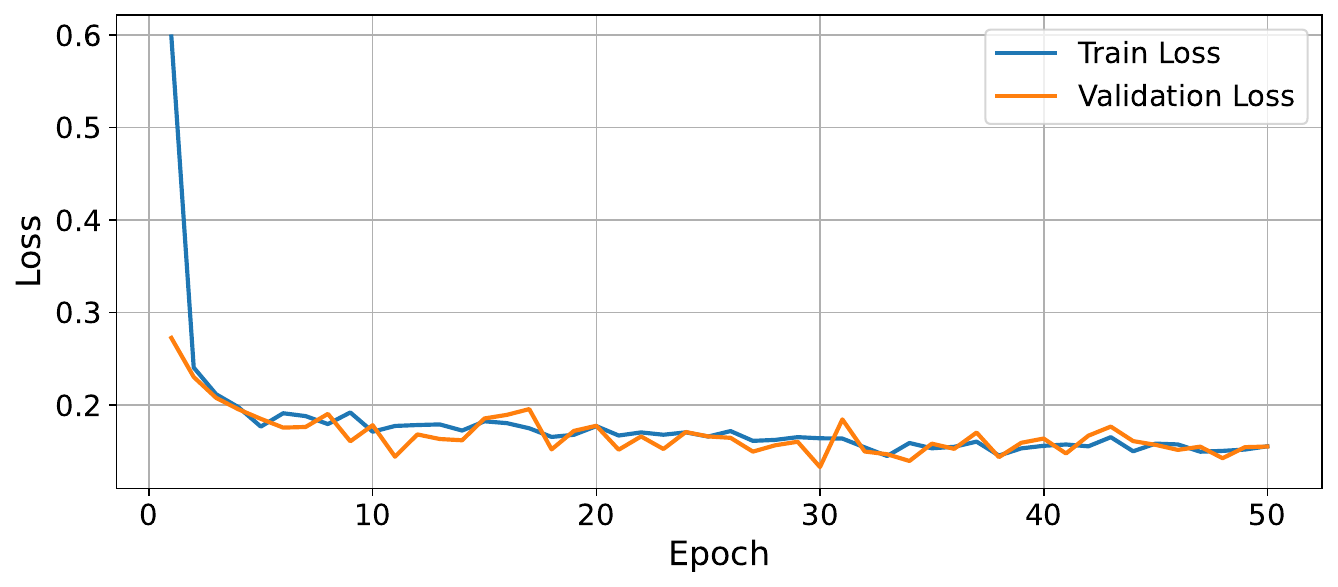}
    \caption{Training and validation loss over 50 epochs.}
    \label{fig: training_loss_curve}
\end{figure}

Early stopping based on validation loss and gradient clipping were used to maintain stable training. The model converged smoothly, as shown in Figure~\ref{fig: training_loss_curve}.

\vspace{-1em}

\subsubsection{Implementation Environment}
All experiments were run on an Apple M2 GPU using the MPS backend. The implementation follows the modular structure of TabDDPM~\cite{tabddpm2023} with the temporal and contextual extensions described in Section~\ref{section: Methodology}.

\subsection{Synthetic Data Generation}
\label{section: synthetic generation}

Following convergence, synthetic sequences were generated using the reverse diffusion process. Starting from Gaussian noise $x_T \sim \mathcal{N}(0, I)$, the model iteratively denoised samples over $T = 1000$ timesteps using the learned function $\epsilon_{\theta}(x_t, t, y, M)$, conditioned on the activity label $y$.

Each reverse step refines the sample by subtracting the predicted noise:

\begin{equation}
x_{t-1} = 
\frac{1}{\sqrt{\alpha_t}}
\left(
    x_t - 
    \frac{\beta_t}{\sqrt{1 - \bar{\alpha}_t}} \,
    \epsilon_\theta(x_t, t, y, M)
\right)
+ \sigma_t z,
\qquad
z \sim \mathcal{N}(0, I),
\end{equation}

where $\{\alpha_t\}$ and $\{\beta_t\}$ follow the cosine variance schedule~\cite{nichol2021improved}.

This iterative reverse process reconstructs noise free sequences $x_0$ for each activity class. Synthetic samples were generated proportionally to the real training distribution, producing a balanced dataset for augmentation (see Section~\ref{section: Results}).

\section{Results}
\label{section: Results}

\vspace{-0.5em}

\subsection{Synthetic Data Generation Results}
\label{subsection: synthetic generation results}

Post convergence, proposed TabDDPM generated 2055 synthetic sequences of length 100 with three sensor channels. Starting from Gaussian noise, the model progressively denoised samples conditioned on activity labels, producing temporally coherent multivariate sequences that mimic the motion dynamics of the original WISDM data.

\begin{figure}[h]
    \centering
    \includegraphics[width=1\linewidth]{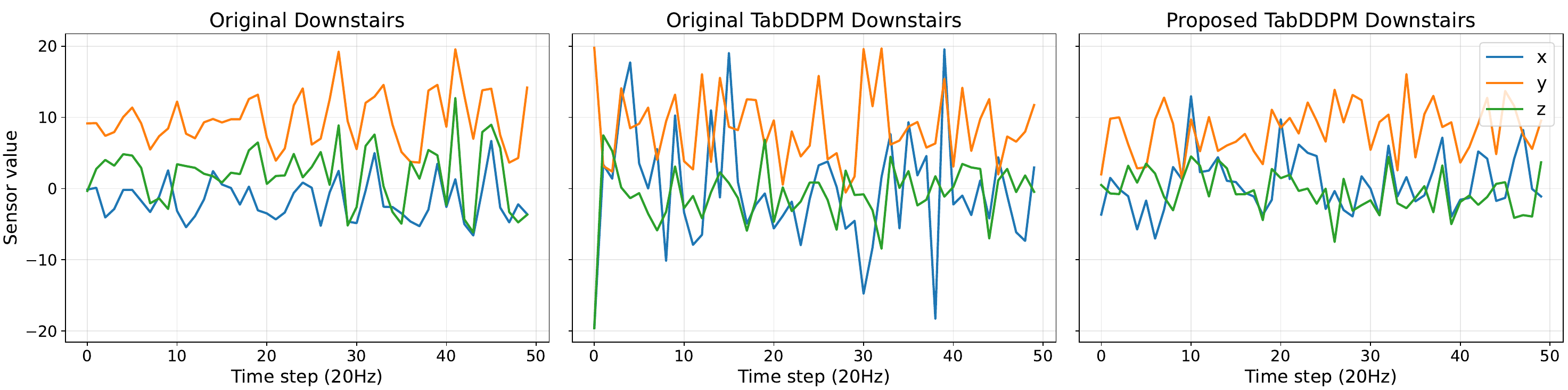}
    \caption{Comparison of real and synthetic accelerometer sequences for the \textit{Downstairs} activity.}
    \label{fig: tabddpm_comparison}
\end{figure}

Figure~\ref{fig: tabddpm_comparison} compares accelerometer sequences for the \textit{Downstairs} activity across real data, the Original TabDDPM, and the proposed method. Proposed TabDDPM produces smoother trajectories with consistent temporal transitions, avoiding the irregular spikes present in the original TabDDPM and demonstrating the benefit of explicit temporal modeling.

Unlike interpolation based approaches such as SMOTE, diffusion methods generate full length sequences; however, proposed TabDDPM exhibits superior temporal coherence, preserving phase relationships critical for downstream classification.

\begin{figure}[h]
    \centering
    \includegraphics[width=1.0\linewidth]{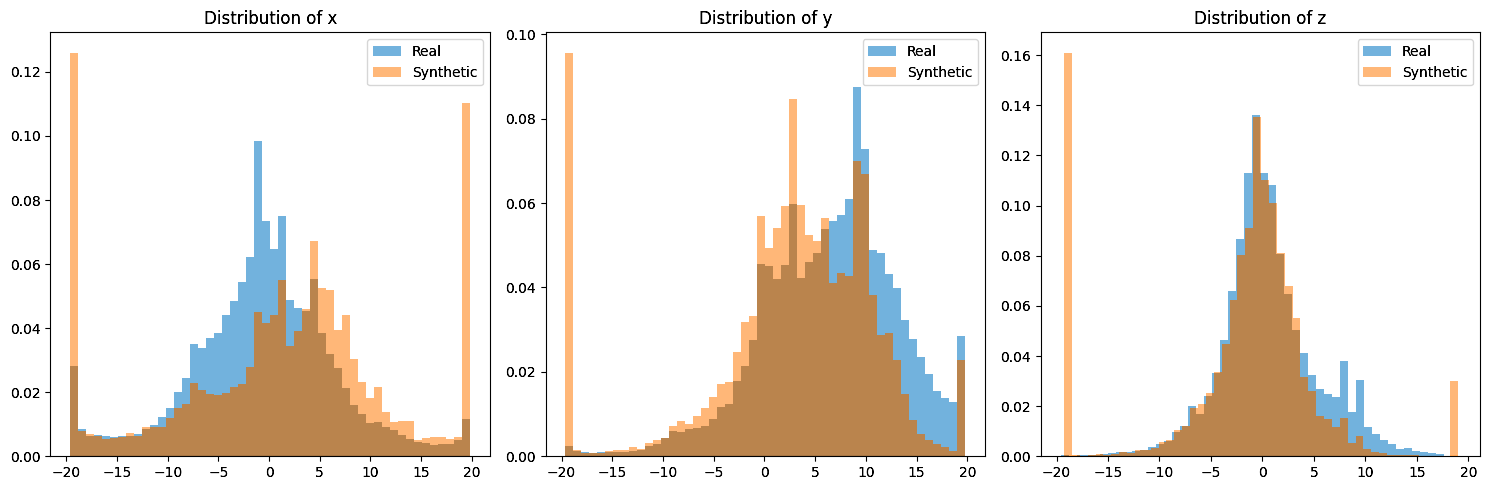}
    \caption{Distribution comparison of accelerometer features ($x$, $y$, $z$) between real and synthetic data.}
    \label{fig: feature_distributions}
\end{figure}

Feature histograms in Figure~\ref{fig: feature_distributions} further confirm strong statistical alignment between real and synthetic sensor distributions, with no evidence of mode collapse or outlier generation.

These findings show that proposed TabDDPM effectively captures latent human motion dynamics and generates realistic, label conditioned synthetic sequences suitable for data augmentation and balancing.

\vspace{-0.5em}

\subsection{Balancing and Data Augmentation}
\label{section: balancing}

\begin{figure}[htbp]
\centering
\includegraphics[width=0.85\linewidth]{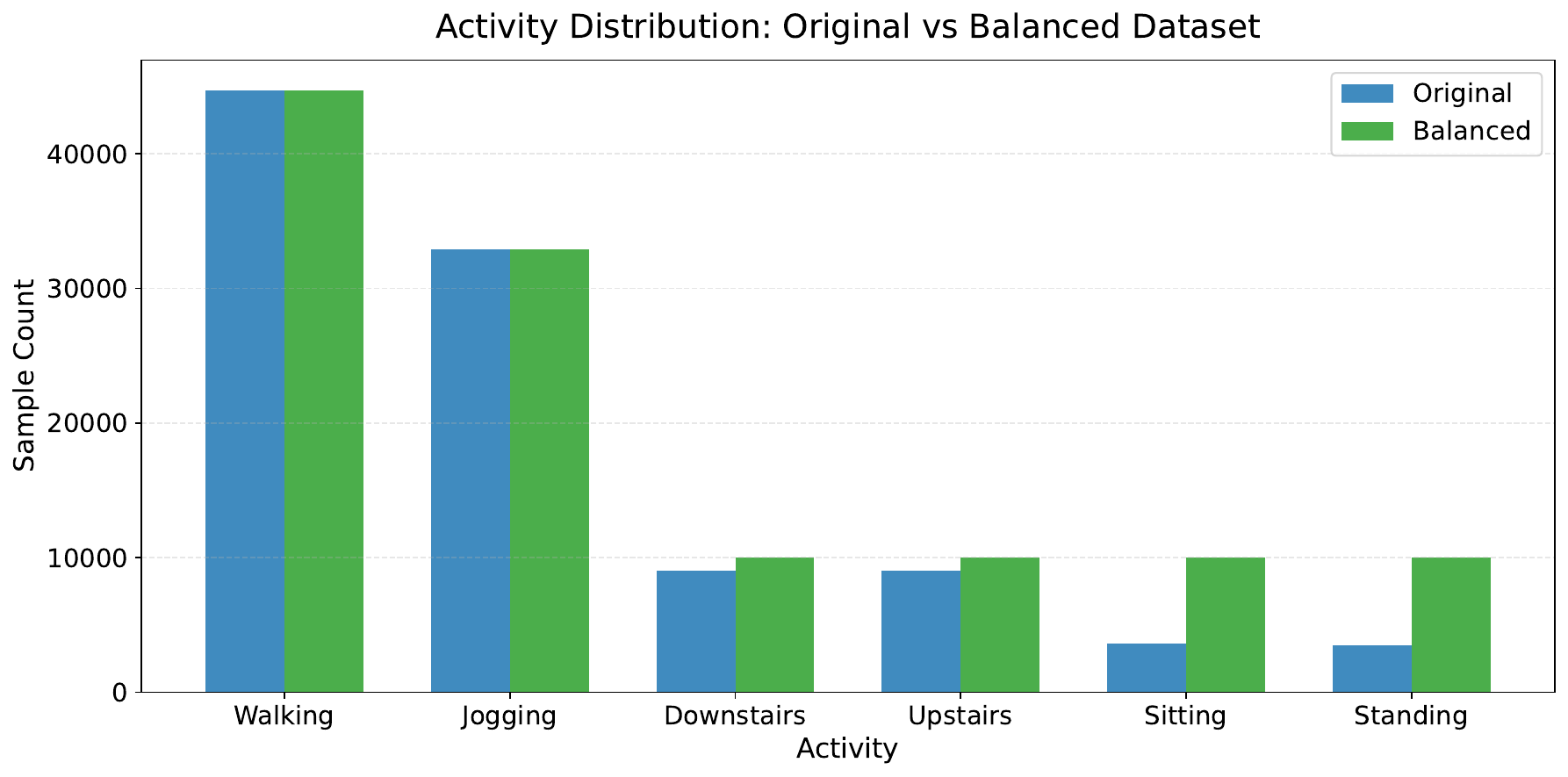}
\caption{Class distribution before and after augmentation.}
\label{fig: balancing_distribution}
\end{figure}

The WISDM dataset shows substantial class imbalance, with locomotive activities overrepresented compared to minority classes (Sitting, Standing, Upstairs, Downstairs). Figure~\ref{fig: balancing_distribution} displays activity wise sample counts before and after augmentation.

\vspace{-0.4em}

To correct this, additional synthetic sequences were generated for the minority classes using SMOTE, Original TabDDPM, and the proposed TabDDPM, upsampling only the underrepresented labels. The resulting balanced datasets were then used to train Random Forest classifiers, as described in the next subsection.

\vspace{-0.5em}

\subsection{Classification Evaluation}
\label{section: classification_evaluation}

To evaluate the effect of generative augmentation on downstream classification, Random Forest classifiers were trained on four versions of the training data: (a) the original unbalanced dataset, (b) the SMOTE balanced dataset, (c) the dataset augmented with synthetic sequences from the Original TabDDPM, and (d) the dataset augmented using the proposed TabDDPM. All models were evaluated on the same held out 20\% partition of real data to ensure a consistent comparison protocol. Random Forest provides a simple, non-temporal baseline for window-level HAR, isolating augmentation effects without confounding temporal modeling.

\begin{quote}
\footnotesize
\textbf{Note:} Hyperparameter configurations were tailored to each method’s structure.
The proposed TabDDPM used sequence inputs of length 100, batch size 64, Conv1D temporal adapters, and conditioning on activity labels and observed/missing masks. Original TabDDPM followed the configuration reported in the original paper~\cite{tabddpm2023}. All classifiers were trained using identical procedures for comparability.
\end{quote}

\subsubsection{Quantitative Results}

Table~\ref{tab:f1_score_comparison} reports overall accuracy and macro F1 score. All methods performed similarly (accuracy $\approx 0.71$, macro F1 $\approx 0.64$), indicating that generative augmentation did not substantially improve majority class recognition.

\begin{table}[h]
\centering
\caption{Overall Accuracy and Macro F1 Score Comparison.}
\label{tab:f1_score_comparison}
\setlength{\tabcolsep}{8pt}
\renewcommand{\arraystretch}{1.2}
\begin{tabular}{l|c|c|c|c}
\hline
\textbf{Method} 
& \textbf{Baseline} 
& \textbf{SMOTE} 
& \begin{tabular}{c}
\textbf{Original} \\[-2pt]
\textbf{TabDDPM}
\end{tabular}
& \begin{tabular}{c}
\textbf{Temporal} \\[-2pt]
\textbf{TabDDPM}
\end{tabular} \\
\hline
Accuracy & 0.71 & 0.71 & 0.70 & \textbf{0.71} \\
Macro F1 & 0.64 & 0.64 & 0.60 & \textbf{0.64} \\
\hline
\end{tabular}
\end{table}

The proposed TabDDPM matches SMOTE/baseline accuracy/F1 but shows temporal advantages (bigram/ACF) and better static class precision (Sitting/Standing). Modest gains reflect WISDM's short-window/minority variability challenges; sequence quality is the primary benefit.

\subsubsection{Confusion Matrix Analysis}

\begin{figure}[h]
    \centering
    \includegraphics[width=0.95\linewidth]{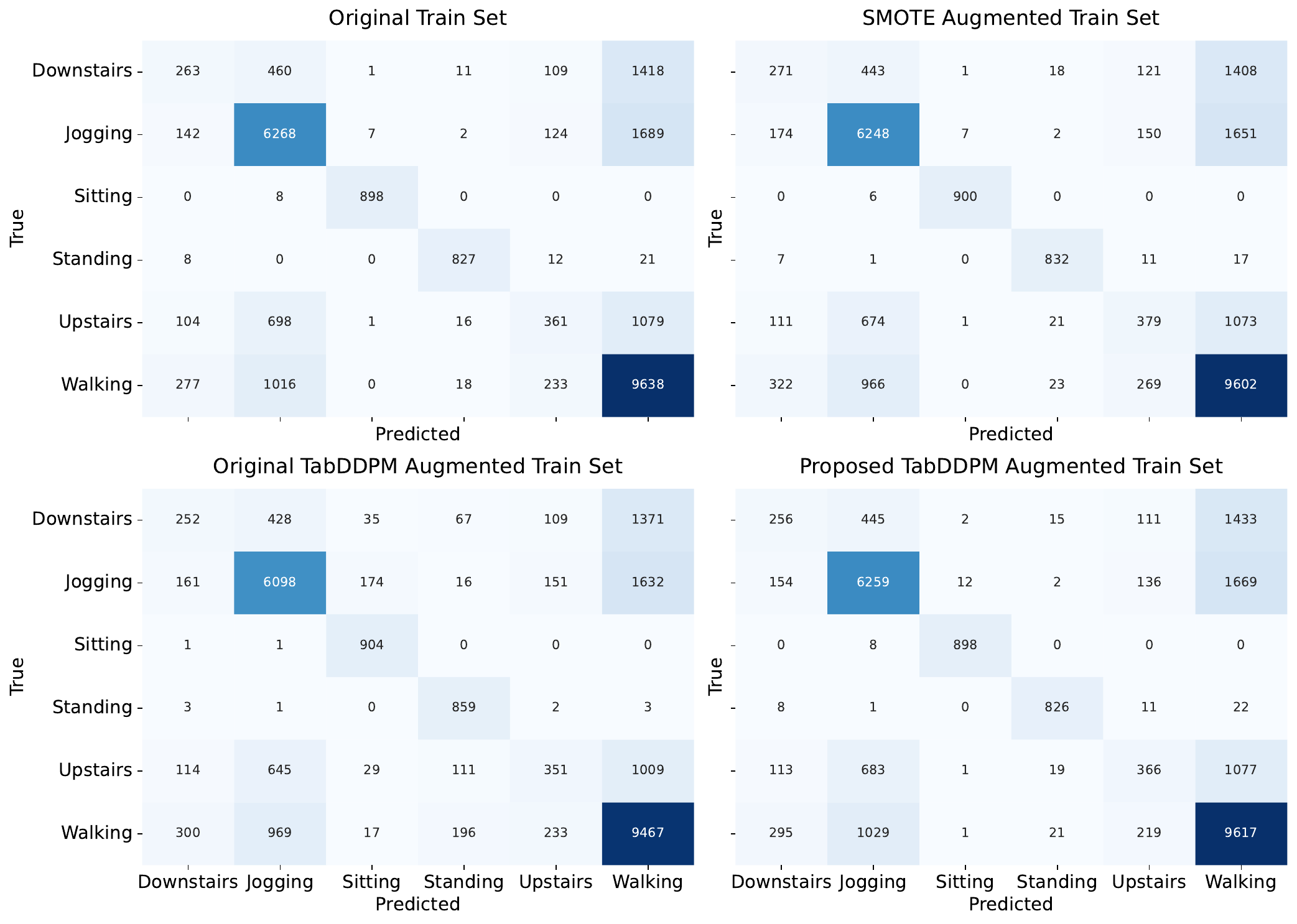}
    \caption{Confusion matrices for Random Forest classifiers trained on (a) Original, (b) SMOTE augmented, (c) Original TabDDPM, and (d) Proposed TabDDPM datasets.}
    \label{fig:confusion_matrix}
\end{figure}

Figure~\ref{fig:confusion_matrix} presents confusion matrices for Random Forest classifiers trained on the original, SMOTE augmented, Original TabDDPM, and proposed TabDDPM datasets. Majority classes (\textit{Walking}, \textit{Jogging}) maintained high performance across all methods, while minority classes showed greater variation.

The Original TabDDPM exhibited reduced precision and recall for static activities (\textit{Sitting}: 0.88, \textit{Standing}: 0.81), reflecting difficulty in generating consistent low variance sequences. Proposed TabDDPM improved these metrics substantially (\textit{Sitting}: 0.99, \textit{Standing}: 0.94), matching SMOTE and the baseline.

Dynamic minority classes (\textit{Downstairs}, \textit{Upstairs}) remained challenging for all approaches, with recall values between 11-16\% due to high intra class variability. Proposed TabDDPM achieved comparable or slightly improved recognition, indicating enhanced temporal stability without increasing confusion with locomotive classes.

Overall, proposed TabDDPM improves temporal coherence and class separability for minority activities while maintaining overall performance similar to interpolation based methods. These results affirm that temporal conditioning in diffusion models yields realistic and coherent synthetic sequences suitable for HAR data augmentation.

\vspace{-0.5em}
\subsection{Temporal Consistency Analysis}

\begin{figure}[htbp]
\centering
\includegraphics[width=0.8\linewidth]{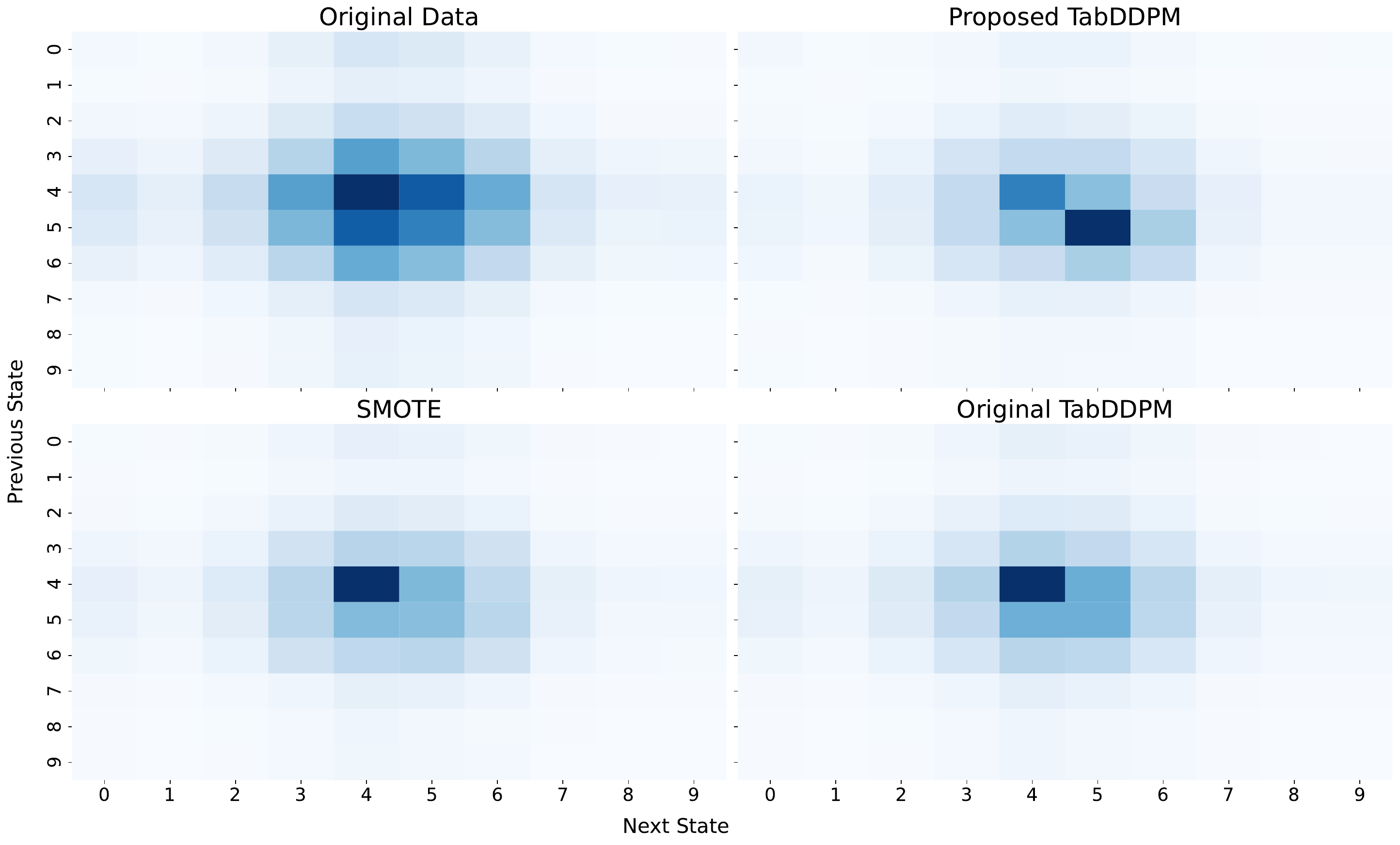}
\caption{Bigram transition matrices for the X-axis showing short-range temporal transitions for (a) Original data, (b) Proposed TabDDPM, (c) SMOTE, and (d) Original TabDDPM.}
\label{fig:bigram}
\end{figure}

To assess temporal coherence, Figure~\ref{fig:bigram} displays bigram transition matrices for the X-axis from original data, SMOTE, Original TabDDPM, and Proposed TabDDPM, reflecting short range sensor transitions. Proposed TabDDPM replicates strong transitions seen in real data, while SMOTE and Original TabDDPM yield flatter, less structured patterns.

\begin{figure}[htbp]
\centering
\includegraphics[width=1\linewidth]{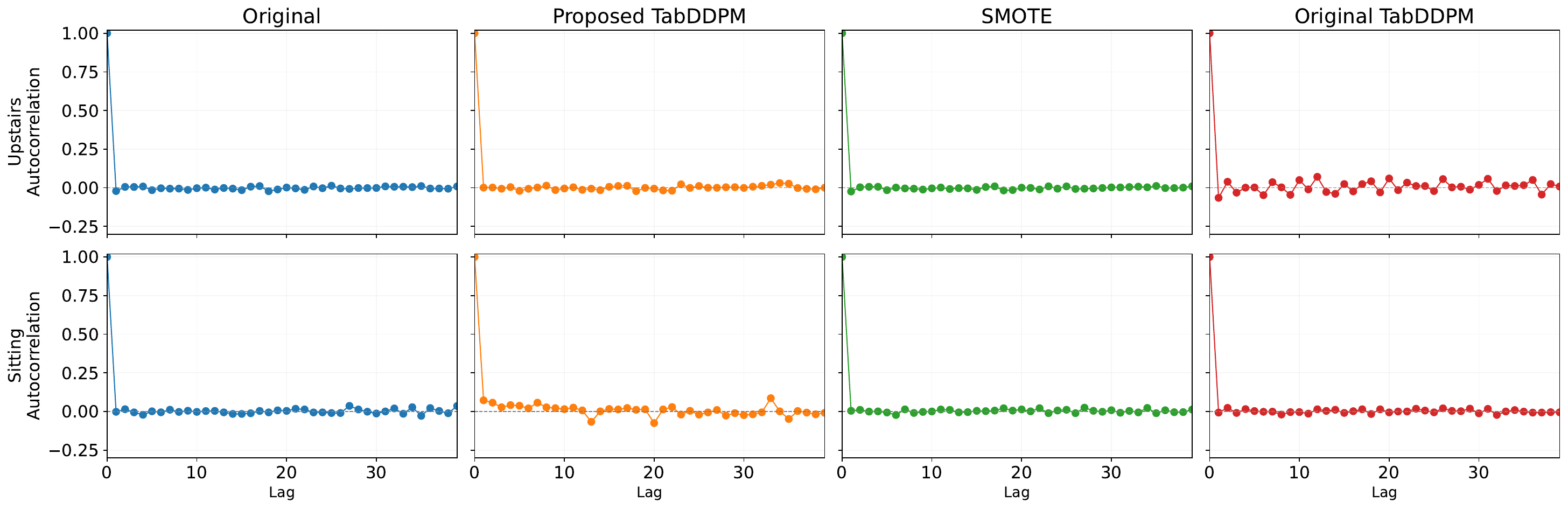}
\caption{Autocorrelation functions (ACF) for \textit{Sitting} and \textit{Upstairs} activities on the Z-axis, comparing temporal dependence across methods.}
\label{fig:acf}
\end{figure}

Figure~\ref{fig:acf} reports autocorrelation functions (ACF) for \textit{Upstairs} and \textit{Sitting} activities over the Z-axis. For the dynamic activity \textit{Upstairs}, the proposed TabDDPM reproduces the short range decay and temporal variability of the real sequence, whereas SMOTE collapses nearly all temporal dependence and Original TabDDPM exhibits irregular oscillations. For the static activity \textit{Sitting}, real data show almost no autocorrelation beyond lag~1. SMOTE matches this flat profile due to interpolation, while Original TabDDPM again produces noise like fluctuations. The proposed TabDDPM displays mild periodic structure, reflecting realistic micro movements (e.g., posture drift or sensor jitter) rather than random noise. These patterns remain low amplitude and more structured than those of the Original TabDDPM, indicating improved stability. Overall, the ACF results demonstrate that the proposed TabDDPM best preserves meaningful temporal dynamics without introducing instability, complementing the bigram findings.

These temporal metrics confirm that Conv1D temporal adapters and context-aware embeddings are the key ingredients: adapters capture local dependencies via lightweight convolutions (Eq.~\ref{eq:adapter_conv}), while timestep/conditional embeddings enable activity-conditioned denoising. On WISDM's 20 Hz data, this preserves short-range stochasticity and periodicity without mode collapse (Fig.~\ref{fig: feature_distributions}). Bad cases like Original TabDDPM's oscillations arise from ignoring cross-timestep relationships, which our extensions mitigate efficiently.

\section{Discussion}
\label{section: Discussion}

The proposed TabDDPM demonstrates that expanding diffusion-based generative models with temporal adapters and conditional embeddings yields multivariate sensor sequences with realistic temporal structure and coherence. This is evidenced by enhanced bigram transition matrices and autocorrelation function (ACF) analyses, both of which show that proposed TabDDPM more faithfully reproduces inter timestep dynamics and preserves the signal distribution of real data. Conv1D temporal adapters and context embeddings are the key ingredients, enabling local dependency capture and activity-conditioned denoising. These adapters enforce local continuity via Conv1D (Eq.~\ref{eq:adapter_conv}), while embeddings/masks enable conditioned, robust denoising.

The primary advantage of this approach lies in temporal quality, not just class-wise accuracy. Compared to Original TabDDPM and SMOTE, the proposed TabDDPM produces smoother, less noisy trajectories for minority activities and preserves natural stochastic and mild periodic dependencies, as confirmed by both bigram and ACF results. Sequence realism improves signal integrity and more accurately corrects class imbalance, yielding synthetic data better suited for downstream tasks.

Although only minor improvements in minority class recognition and equivalent overall classification accuracy versus SMOTE and baseline Random Forest are observed, the superior temporal consistency and diversity of synthetic data underscore the need for models and evaluation metrics that prioritize time-series quality. Traditional classifiers like Random Forest cannot fully exploit the richer temporal structure, explaining modest accuracy gains despite superior bigram/ACF. Temporal models (LSTM/CNN) would better leverage this, planned for future work along with advanced metrics.

Finally, limitations in activity class separability especially for \textit{Upstairs} and \textit{Downstairs}, highlight constraints of the WISDM dataset’s triaxial accelerometer setup. A key limitation is evaluation on a single dataset (WISDM), limiting robustness claims across HAR benchmarks. Future work will validate on additional datasets. Continued development of transformer based temporal diffusion architectures and inclusion of multimodal sensor streams will be critical for further gains in HAR and synthetic data realism.

\section{Conclusion and Future Directions}
\label{section: Conclusion and Future Directions}

This paper introduced Temporal TabDDPM, a diffusion based generative model leveraging Conv1D temporal adapters and context-aware conditioning to synthesize realistic, temporally coherent sensor sequences for human activity recognition. By reformulating data into windowed sequences and integrating timestep, conditional, and mask embeddings, the model captures essential temporal dynamics and improves minority class balance.

On the WISDM dataset, proposed TabDDPM matched baseline and SMOTE in classification accuracy (0.71) and macro F1-score (0.64), but demonstrated notable gains in temporal consistency validated by bigram transition and autocorrelation analyses which set it apart from interpolation and non temporal diffusion approaches. The lightweight adapter architecture offers sequence awareness with low computational cost, highlighting the benefits of targeted convolutional temporal modeling.

Despite improvements in sequence realism and minority class representation, classifier gains remain modest, evaluation on a single dataset (WISDM) also limits generalizability, indicating the need for richer input signals or specialized temporal evaluation metrics. Future work will expand this framework to multimodal HAR data, explore transformer-based temporal diffusion models, consider advanced temporal metrics, and examine generalization across additional HAR datasets, temporal classifiers, domains, and longer sequence lengths to further advance generative fidelity and utility.

\end{document}